\documentclass{bmvc2k}
\usepackage{graphicx} 
\usepackage{subfigure} 
\usepackage{tikz}
\usepackage{epsfig}

\usepackage{float}

\usepackage{amsthm}
\usepackage[export]{adjustbox}
\usepackage{tabularx}
\usepackage{float}
\usepackage{wrapfig}
\usepackage{booktabs}
\usepackage{epstopdf}
\usepackage{verbatim}

\usepackage{algorithm}
\usepackage{algorithmic}
\usepackage{amsmath}

\usepackage{amssymb}

\title{Dynamic Steerable Blocks \\ in Deep Residual Networks}

\addauthor{J\"{o}rn-Henrik Jacobsen}{j.jacobsen@uva.nl}{1}
\addauthor{Bert de Brabandere}{bert.debrabandere@esat.kuleuven.be}{2}
\addauthor{Arnold W.M. Smeulders}{A.W.M.Smeulders@uva.nl}{1}

\addinstitution{
 Informatics Institute\\
 University of Amsterdam \\
 Amsterdam, NL
}
\addinstitution{
 ESAT-PSI \\
 KU Leuven,\\
 Leuven, BE
}

\runninghead{Jacobsen, de Brabandere, Smeulders}{Dynamic Steerable Blocks}

\begin{document}

\newcommand{\softmax}{\rm smax}
\newcommand{\rb}{\rangle}
\newcommand{\lb}{\langle}
\newcommand{\om}{\omega}
\newcommand{\la}{\lambda}
\newcommand{\Real} {{\rm Real}}
\newcommand{\R} {{\mathbb R}}
\newcommand{\E} {{\mathbb E}}
\newcommand{\N} {{\mathbb N}}
\newcommand {\h} {{\mathfrak h}}
\newcommand{\Z} {{\mathbb Z}}
\newcommand{\C} {{\mathbb C}}
\newcommand{\HH} {{\cal H}}
\newcommand{\Ld} {{\bf L^2}}

\newtheorem{theorem}{Theorem}[section]
\newtheorem{proposition}{Proposition}[section]
\newtheorem{corollary}{Corollary}[section]

\maketitle

\begin{abstract}
Filters in convolutional networks are typically parameterized in a pixel basis, that does not take prior knowledge about the visual world into account. We investigate the generalized notion of frames designed with image properties in mind, as alternatives to this parametrization. We show that frame-based ResNets and Densenets can improve performance on Cifar-10+ consistently, while having additional pleasant properties like steerability. By exploiting these transformation properties explicitly, we arrive at dynamic steerable blocks. They are an extension of residual blocks, that are able to seamlessly transform filters under pre-defined transformations, conditioned on the input at training and inference time. Dynamic steerable blocks learn the degree of invariance from data and locally adapt filters, allowing them to apply a different geometrical variant of the same filter to each location of the feature map. When evaluated on the Berkeley Segmentation contour detection dataset, our approach outperforms all competing approaches that do not utilize pre-training. Our results highlight the benefits of image-based regularization to deep networks. 
\end{abstract}


\begin{figure}
\includegraphics[width=1.\textwidth, center]{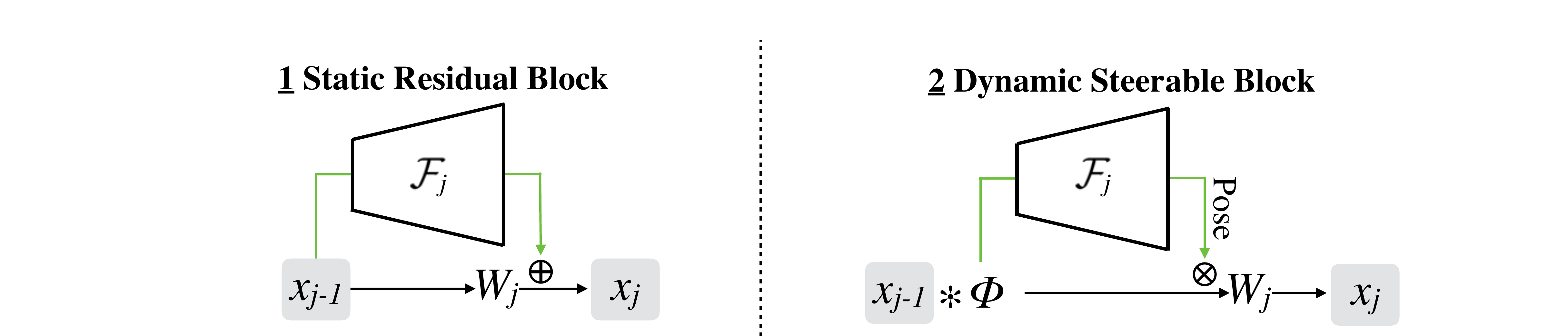}
\caption{Left is a classical residual block as used in vanilla ResNets~\cite{he2016deep}. Outputs $x_{j-1}$ of the previous block are combined additively with the output of a small stack of convolutional layers $\mathcal{F}_j$ to form the final output $x_j$. We augment this formulation in two ways to achieve our dynamic block formulation on the right. First, we apply a change to a steerable frame $\Phi$ on the input $x_{j-1}$ and second we replace the addition operation with a multiplication. This permits us to interpret $\mathcal{F}_j$ as a pose estimating network, that directly outputs the linear projection coefficients, transforming the basis and subsequently the effective filters adaptively in a dynamic and, if desired, location dependent manner.
\label{fig:brain}}
\end{figure}
\section{Introduction}
Deep Convolutional Neural Networks (CNNs) are the state-of-the-art solution to many vision tasks~\cite{lecunNature15deep}. However, they are known to be data-inefficient, as they require up to millions of training samples to achieve their powerful performance~\cite{zeiler2014visualizing,zhou2014places}. In this work, we propose a formulation of CNNs that more efficiently learns to exploit generic regularities known to be present in the data a priori.

For images, as well as any other sensory data, CNNs typically learn filters from individual pixel values. In this paper, we show that alternatives to the pixel basis are more natural formulations for learning models on locally well-understood data like images. We show increased classification performance in state-of-the-art ResNets~\citep{he2016identity} and Densenets~\citep{huang2016densely} on the highly competitive Cifar-10 classification task by replacing the pixel basis with a basis more suitable for natural images. Further, we show that such a replacement naturally leads to powerful extensions of residual blocks as dynamic steerable interpolators that can steer their filters conditioned on the input with respect to pre-defined continuous geometric transformations like rotations or scalings. The proposed block allows the network to adaptively learn the degree of local invariance required for each filter, by decoupling filter learning and local geometrical pose adaption. We show the effectiveness of this approach on the BSD500 boundary detection task, where precise and adaptive local invariances are key. We outperform all competing non-pretrained methods with our approach. \\

Our contributions:  
\begin{itemize}
\item We introduce the notion of frame bases to CNNs and show that classically used frames from Computer Vision aid optimization, when compared to the commonly used pixel basis. We illustrate this by improving classification performance on Cifar-10+ for multiple ResNet and Densenet architectures, by mere substitution of the pixel basis with a frame only. 
\item Exploiting the steerability properties of frames further, we derive Dynamic Steerable Blocks that are able to continuously transform features in a locally adaptive manner and illustrate the approach on a synthetic tasks to highlight the advantages over competing approaches.
\item To evaluate the practicality of our proposed approach, we apply Dynamic Steerable Block networks on the BSDS-500 contour detection task~\citep{bsds500} where we achieve increased performance among competing methods that do not utilize pre-training.
\end{itemize}

The paper is organized as follows. First, we review related literature of alternative parametrizations and incorporation of prior geometrical knowledge into CNNs. Secondly, we introduce the theoretical framework of frame-based CNNs and steerable two-factor blocks, which our work rests upon. Third, we show that careful reparametrization of CNNs can increase performance on natural image classification. Lastly, we show how to extend frame-based Resnet blocks to dynamic steerable blocks that have the ability to dynamically adapt filters, with several advantages and promising results on the Berkeley Contour Detection dataset.

\section{Related Work}

Convolutional Networks with alternative bases have been proposed with various degrees of flexibility. A number of works utilizes change of basis to stabilize training and increase convergence behavior~\cite{rippel2015spectral,arjovsky2015unitary}. Another line of research is concerned with complex-valued CNNs, either learned~\cite{tygert2016mathematical}, or fully designed like the Scattering networks~\cite{bruna2013invariant,oyallon2015deep}. 

Scattering, as well as the complex-valued networks, rest upon a direct connection between the signal processing literature and CNNs. Inspired by the former, Structured Receptive Field Networks are learned from an overcomplete multi-scale frame, effectively improving performance for small datasets due to restricted feature spaces~\cite{Jacobsen2016}. Closely related is another line of recent promising work on group-equivariant~\cite{cohen2016group,dieleman2016exploiting} and steerable CNNs~\cite{worrall2016harmonic,cohen2016steerable}. The latter build steerable representations via appropriately chosen basis functions, illustrating that CNNs with well-chosen geometrical inductive biases consistently outperform state-of-the-art approaches in multiple domains. However, both all these approaches rely on hand-engineered types of representations and none considers locally adaptive filtering with learned degree of invariance, we aim to bridge this gap. Inspired by CNNs learned from alternative bases, we introduce the general principle of Frame-based convolutional networks that allow for non-orthogonal, overcomplete and steerable feature spaces.

Steerable frames are a concept established early for signal processing. Initially introduced by~\cite{freeman1991design}, the concept was extended to the Steerable Pyramid by~\cite{simoncelli1995steerable} and to a Lie-group formulation by~\cite{hel1998canonical,michaelis1995lie}. Further, steerability has recently been related to tight frames, presenting Simoncelli's Steerable Pyramid and multiple other Wavelets arising as a special case of the non-orthogonal Riesz transform~\cite{unser2013unifying}. Steerable pyramids have been applied to CNNs as a pre-processing step~\cite{xue2016visual}, but have not yet been learnable. We incorporate steerable frames in CNNs to increase their de facto expressiveness and to allow them to learn their configurations, rather than picking them a priori. 

Another way to impose structure onto CNN representations and subsequently increase their data-efficiency is by pre-defining the possible transformations, as done in Transforming Autoencoders~\cite{hinton2011transforming}, which map their inputs from the image to pose space through a neural network. The Spatial Transformer Networks~\cite{jaderberg2015spatial} learn global, and deformable convolutional networks~\cite{dai2017deformable} local transformation parameters in a similar way while applying them to a nonlinear co-registration of the feature stack to some estimated pose. Dynamic Filter Networks~\cite{de2016dynamic} move one step further and estimate filters for each location, conditioned on their input. These approaches are all dynamic in a sense that they condition their parameters on the input appearance. Our proposed dynamic residual block can be interpreted as a middle ground that combines the idea of Dynamic Filter Networks with explicit pose prediction into blocks that can locally estimate filter poses from a continuous input space. As such, we overcome the difficulty of estimating local filter pose, while being able to separate pose and feature learning globally without the need for differentiable samplers or locally connected layers.

\section{CNN Bases Beyond Pixels} 

The most general set of viable bases to learn filters from are called \textit{Frames}~\cite{christensen2003introduction}. Frames are a natural generalization of orthogonal bases and are spanning sets that span the same space of functions an orthogonal basis does, while allowing for overcomplete representations and hence more densely sampled parameter spaces. 

\begin{figure}
\includegraphics[width=.45\textwidth, center]{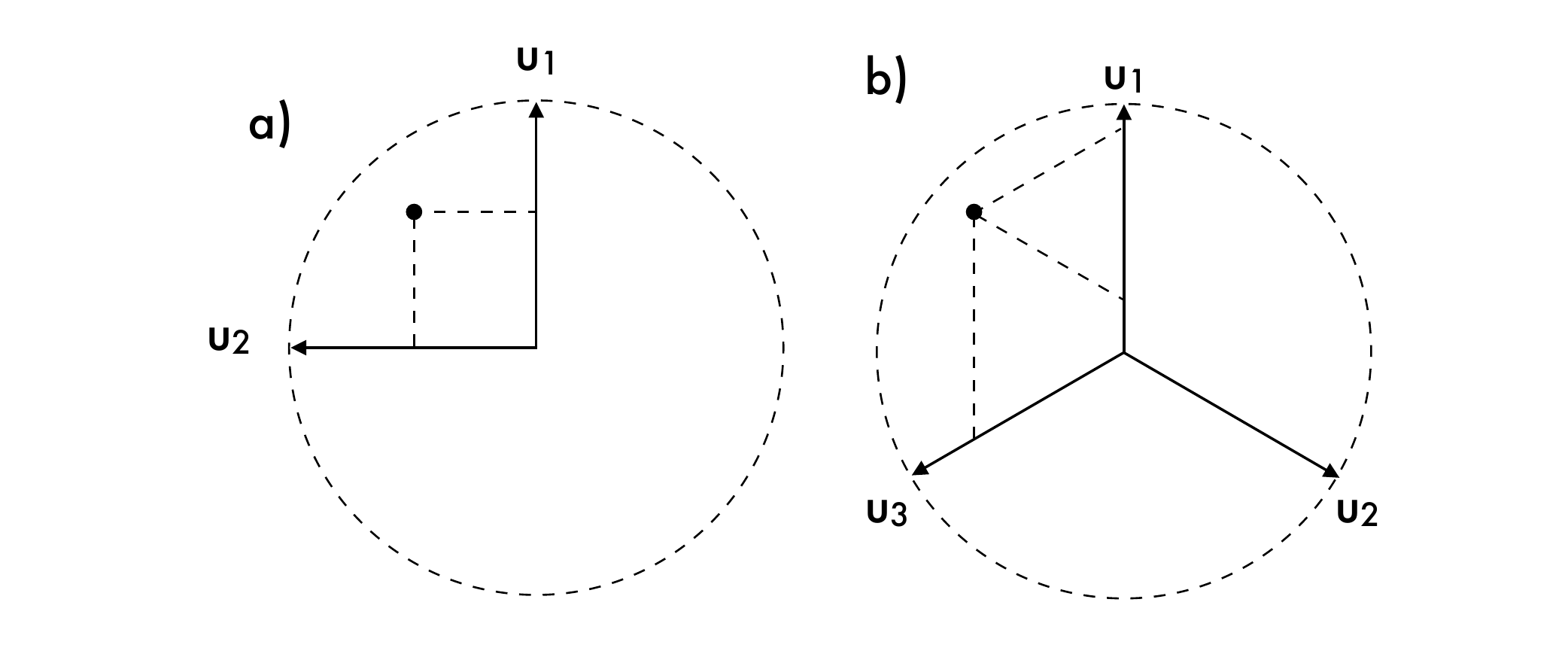}
\vspace*{-5mm}
\caption{a) Is an orthonormal basis in $\mathbb{R}^2$, $u_1$ and $u_2$ are linearly independent and span the space of $\mathbb{R}^2$. A dot in this example represents a filter in a convolutional network with coefficients $\{u_1,u_2\}$. b) A tight frame in $\mathbb{R}^2$. $u_1$, $u_2$  and $u_3$ are linearly dependent. A dot in this example represents a convolutional filter with coefficients $\{u_1,u_2,u_3\}$. The frame is an overcomplete representation, again spanning $\mathbb{R}^2$ and again preserving the norm. Note that the set of filter coefficients as represented by the dot is not unique. Thus even if one $u$ is obstructed by noisy updates or measurements, the filter may still be robust. 
\label{fig:frame_explain}}
\end{figure}

Frames can be seen as a superset of orthogonal bases in the sense that every basis is a frame, but not the reverse, see figure~\ref{fig:frame_explain}. Frames have three main advantages: 1) Frames can spell out signal properties more explicitly, facilitating optimization when good frames for the type of data are known, as is the case for many types of signals like images or video~\cite{christensen2003introduction}. 2) Frames allow for overcomplete representations of the signal adding robustness and regularization to the optimization procedure, as they are more stable when measurements or updates are noisy. 3) Many frames are steerable and thus provide us with a signal representation that can be dynamically steered by simple linear projections, i.e. they have the ability to linearize group actions. Properties 1) and 2) are illustrated in experiment 3.1, where we merely replace standard pixel bases with alternative frames and evaluate the effect. Property 3) is illustrated in experiments 3.2, where we leverage the steerability property of some frames explicitly.

\begin{figure}
\includegraphics[width=.5\textwidth, center]{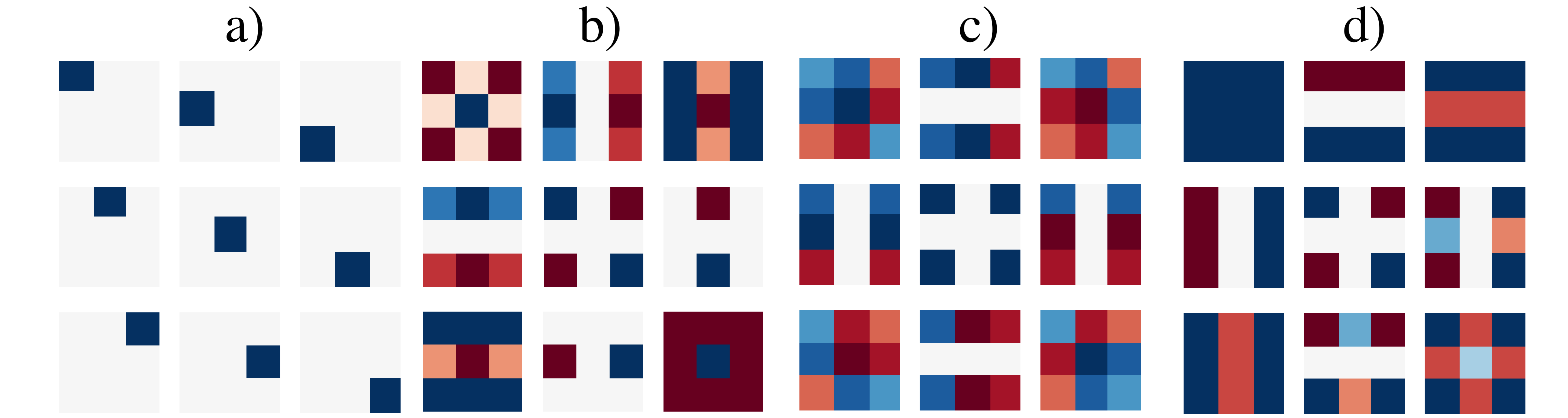}
\vspace*{-3mm}
\caption{An illustrative plot of multiple 3x3 spanning sets $\Phi$: a) Pixel-basis, b) Gaussian Derivatives (first 9 atoms), c) Non-orthogonal Framelet, d) Naive Frame. Note the increased symmetries in the three latter.
\label{fig:frame_examples}}
\vspace*{-5mm}
\end{figure}

In a standard convolutional network, a filter kernel is a linear combination over the standard (pixel) basis for $l^2(\mathbb{N})$. This pixel basis is composed of delta functions for every dimension and $W_i$ is the $i_{th}$ filter of the network with parameters $w^i_n$. Without loss of generalization the orthonormal standard basis can be replaced by a frame to include steerability, non-orthogonality, overcompleteness and increased symmetries into the representation.
Changing from the pixel to an arbitrary frame is as simple as replacing the pixel basis $e_n$ with a frame of choice with elements $\phi_m$ as follows:
\begin{equation}
W_j \Phi(x_{j-1}) = \sum_{n=1}^{N} w_n^j e_n \star x_{j-1} \equiv \sum_{m=1}^{M} w_m^j \phi_m \star x_{j-1},
\label{eq:frame}
\end{equation}
where  $w^i_1,...,w^i_m$ are again the filter coefficients being learned. Note that after optimization with an overcomplete frame is done, the resulting network can be rewritten in terms of the standard pixel basis. Therefore frames only act as a regularizer during training, they do not increase the effective parameter cost of the network. In practice for CNNs working on images we investigate four typical choices of frames: i) the vanilla orthogonal pixel basis, ii) Gaussian derivatives, one of the most widely used overcomplete frames from the computer vision literature (also used in SIFT)~\cite{lowe1999object, florackIVC92scaleAndDiffStruct, koenderinkBioCyb87localGeoInVisSys, Jacobsen2016}, iii) Framelets, a non-orthogonal but not overcomplete basis, designed for images~\cite{daubechies2003framelets} and iv) A "naive" frame of the form $x^p y^q$ derived from steerability requirements~\cite{hel1998canonical} but with no image properties in mind. See figure ~\ref{fig:frame_examples} for plots of these frames and experiments in section \ref{frames_cifar} for their performance. \\

Many frames are steerable, which means equivariant with respect to some group of transformation, they linearize the action of these transformation groups. More formally, if $W$ are the weights of a filter, $g(\tau) \in G$ is some transformation of the input $x$ and the basis $\Phi$ is steerable under this transformation, one can find a linear mapping $\mathcal{F}(\tau)$, such that:
\begin{equation}
W \mathcal{F}(\tau)  \Phi(x_{j-1}) = W \Phi( g(\tau) x_{j-1}).
\label{eq:steer}
\end{equation}
This means instead of steering the effective filters represented by $W  \Phi(x_{j-1}) $, it is sufficient to steer the basis $ \Phi$ and the effective filters will be transformed accordingly.
Thus, a change to a steerable basis $\Phi$, makes it possible to dynamically adapt filters via linear projections, while also opening new ways to learn local and global invariants efficiently. We will show below, that this property permits to dynamically transform filters and to directly learn the degree of invariance to variabilities like rotation, scaling and others, depending on the type of basis used. 

Steerability allows to separate a feature's pose from its canonical appearance. From equation \ref{eq:frame} follows that a steerable version of an arbitrary filter $W_i(x,y)$ under a k-parameter Lie group can be expressed as:
\begin{equation}
g(\tau)W_i(x,y)=\sum_{n=1}^{N} w^i_n g(\tau) \phi^i_n.
\end{equation}
And by substituting according to equation \ref{eq:steer} it follows:
\begin{equation}
g(\tau)W_i(x,y)=\sum_{n=1}^{N} w^i_n \sum_{m=1}^{M} \mathcal{F}_m(\tau) \phi_m(x).
\label{eq:steer_feature}
\end{equation}
Thus it is sufficient to determine the group action on the fixed frame by steering it to separate the canonical feature itself from its k-parameter variants, i.e. $\phi^i_n$ are the frame coefficients underlying each feature $W_i(x,y)$ and $\mathcal{F}(\tau)_m$ are the steering functions governing the transformation of $g(\tau)$ acting on $W_i(x,y)$ as a whole. In the following section, we will connect this insight to frame-based static residual blocks and turn them into dynamic two-factor blocks, that perform geometrically regularized locally adaptive filtering.

To achieve precise geometrical regularization, one can further derive the steering equations for the particular steerable frame at hand and use the resulting trigonometric functions as activation functions, which is suitable for learning in a CNN. While these activation functions can be omitted in more general tasks like classification where the demand on local transformations is not precisely defined, we will show below that they serve as important regularizers in tasks where precise geometric adaption is needed, as examplified in the boundary detection experiments. For brevity, we moved steering equation derivation and steerability proofs to the supplementary material.

\subsection{Dynamic Steerable Two-Factor Blocks}

Deep Residual Networks (ResNets) are among the best performing current approaches to convolutional networks. Instead of optimizing single layers, they consist of convolutional blocks with skip connections, where the output of block $j$ is defined as $x_j= \mathcal{H}(x_{j-1})= \mathcal{F}(x_{j-1})+x_{j-1}$. $ \mathcal{F}$ is typically a stack of convolution layers, batch normalizations~\cite{ioffe2015batch} and nonlinearities~\cite{he2016deep, he2016identity}. Such residual blocks overcome vanishing gradients and facilitate optimization~\cite{greff2016highway}, leading to very simple, but very deep networks with no need for pooling and fully connected layers. In the following sections, we introduce extensions of residual blocks as two-factor models, that overcome the static nature of typical CNNs by leveraging previously introduced steerable frames. A simple extension transforms static residual blocks into dynamic modules, able to change the geometrical pose of their filters conditioned on the input during training \textit{and} inference time in a locally adaptive fashion.

The key perspective the steerable two-factor block relies upon is to change the residual block from being an additive model of the form:
\begin{equation}
\mathcal{H}(x)= \mathcal{F}(x)+W x,
\end{equation}
where W is the shortcut projection introduced in a recent improvement~\cite{he2016identity}, to a multiplicative two-factor model of the form:
\begin{equation}
\mathcal{H}(x)= W  \mathcal{F}(\Phi(x)) \circ \Phi(x).
\label{eq:two-factor}
\end{equation}
Here, the factor $W \Phi(x)$ represents a canonical feature, while $\mathcal{F}(\Phi(x))$ represents its pose. The function $\mathcal{F}$ estimates the local pose. The space of possible poses can be pre-conditioned by choosing a suitable frame $\Phi$, that are steerable under pre-defined sets of deformations and thus span an invariant subspace of all transformed versions of the basis itself. 

The advantage of our proposed dynamic steerable block over common methods achieving local invariance is that our method goes beyond maximum search in local pose estimation and thus overcomes the inherent ambiguity of local maximum search in many natural images. This is achieved by allowing $\mathcal{F}$ to include context around the current filter location into the estimation, enabling stable and task-dependent pose estimation. Figure \ref{fig:blobs} illustrates an instance of this problem where the maximum response can not solve the task and a learned pose estimation finding poses that are neither maxima nor minima, is necessary. Note however, that if invariance is harmful for the task at hand, our model can also learn to fall back to standard CNN representations as well.

The blocks used in the boundary detection experiments are based on Gaussian derivatives, steerable with respect to continuous rotations and small ranges of isotropic scalings ($\sigma=$0.8-1.5). The Dynamic Two-Factor Block consists of four processing steps. 1) Change from input to frame space by convolving with frame, 2) The interpolator network $\mathcal{F}(\Phi(x))$ estimates local pose from this invariant subspace, outputting a set of pose variables for each location in the image and for each input/output channel (this can be changed depending on application). For the interpolator network $\mathcal{F}$ we used a small network with 8-16 units and three layers with tanh nonlinearities as they seemed suitable to approximate trigonometric steering equation solutions. For scalings, we found softplus and relu nonlinearities to work well. Optional: 3) the steering functions derived in section 2.4 are applied to the pose variable maps and effectively act as nonlinear pose-parametrized activation functions that regularize the interpolation network to output an explicitly interpretable pose space. 

4) A 1x1 convolution layer is applied to the already transformed frame outputs, this convolution represents the weights $w^i_j$ of each feature governing the canonical appearance of the $i_{th}$ feature map in the $j_{th}$ layer. 
\section{Experiments}
The experimental section is organized in two parts. Experiment \ref{frames_cifar} illustrates that mere replacement of the pixel basis with frames outperforms highly optimized and flexible approaches like ResNets and Densenets, in domains where they excel, given a good frame for the data at hand is known. Note that this part does not explicitly make use of steerability, but focuses on illustrating the effect of various frame choices according to equation \ref{eq:frame}. Experiment \ref{segment} makes explicit use of steerability, according to equation \ref{eq:steer_feature} \& \ref{eq:two-factor}, illustrating our dynamic steerable block mechanism and applying it on a difficult real-world task of boundary detection, where we outperform all other non pre-trained methods, among which one is rotation invariant.
\subsection{Generalized Bases on Cifar-10+}
\label{frames_cifar}

\begin{table}
\begin{tabular}{@{}lccccc@{}} 
Method &  ResNet110 & ResNet164 & DensenetK12L40 & DensenetK12L100 \\ \midrule
Pixel Basis &  6.70$\pm$ 0.16\% & 5.88$\pm$ 0.22\%  & 5.26$\pm$ 0.19\% & 4.16 $\pm$ 0.18\%  \\
Image Frame &  \textbf{6.11}$\pm$ 0.19\% &  \textbf{5.33}$\pm$ 0.15\%  &  \textbf{4.97}$\pm$ 0.18\% &  \textbf{3.78} $\pm$ 0.17\%  \\
Naive Frame &  7.29$\pm$ 0.31\% & 6.93$\pm$ 0.29\%  & 6.40$\pm$ 0.21\% & 5.25 $\pm$ 0.20\%  \\
\bottomrule
\end{tabular}
\vspace*{+5mm}
\caption{Results of 3 frames on Cifar-10+, reported as average over 5 runs with standard deviation. We compare models with a standard pixel-basis, a steerable frame basis designed for natural images and a naive steerable frame as an example of a frame that does not take natural image statistics into account. The natural image statistics based frame outperforms the pixel-basis consistently, while the naive frame consistently performs about 1\% worse than the baseline, highlighting the benefit of a frame suitable for the type of input data.}
\label{tab:frames}
\end{table}

To show the effect of replacing the commonly used pixel-basis with frames, we compare different frames in multiple state-of-the-art deep Residual Network~\cite{he2016identity} and Densenet~\cite{huang2016densely} architectures on the Cifar-10~\cite{krizhevsky2009learning} dataset with moderate data augmentation of crops and flips. We evaluated our approach on two different network sizes that still comfortably train on one GPU over night each. The setup used for the ResNet is as described in~\cite{he2016identity}. The batch size is chosen to be 64 and we train for 164 epochs with the described learning rate decrease. The ResNet architectures used are without bottlenecks having 56 and 110 layers. For the Densenets we follow~\cite{huang2016densely} and evaluate on the K=12 and L=40, and the K=12 and L=100 models. We run our experiments in Keras~\cite{chollet2015keras} and Tensorflow~\cite{abadi2016tensorflow}. In the first experiment, we run the models on the standard pixel basis to get a viable baseline. Results are in line or better with the numbers reported by the authors. Secondly, we replace the pixel-basis with widely-used frames that take natural image statistics into account, namely non-orthogonal, overcomplete Gaussian derivatives~\cite{florackIVC92scaleAndDiffStruct} and non-orthogonal framelets~\cite{daubechies2003framelets} in an alternating fashion, yielding superior performance compared to the pixel-basis by replacement only. We also show that the naively derived  $x^p y^q$ frame performs consistently worse than the other two choices, as it does not take natural image properties into account. 

The frames used, are shown in figure ~\ref{fig:frame_examples}. The results are reported in table \ref{tab:frames}. The fact that the pixel-basis can be replaced by steerable frames with well-understood properties while performance improves is remarkable. Frame-based CNNs run at approximately the same runtime as vanilla CNNs.

\subsection{Dynamic Steerable Blocks for Boundary Detection}
\label{segment}

We evaluate our proposed dynamic steerable two-factor blocks on boundary detection, a natural task for locally adaptive filtering. In the first part, we illustrate properties of the proposed mechanism and in the second part we apply it to a challenging real-world problem, where we outperform competing approaches.
\subsubsection{Evaluating Boundary Detection Properties on Textured Blobs}
\begin{figure}[H]
\vspace*{-5mm}
\includegraphics[width=.9\textwidth, center]{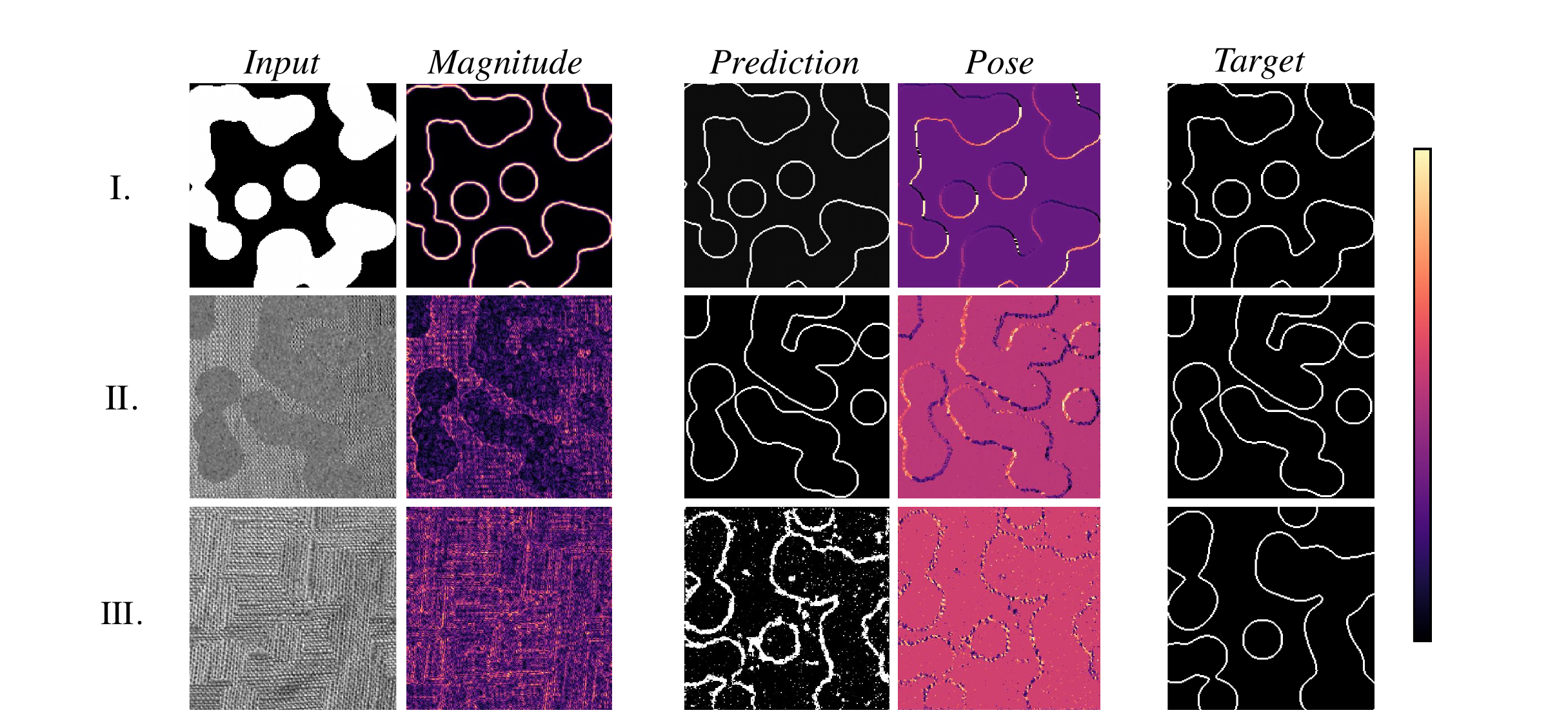}
\vspace*{-3mm}
\caption{Results of boundary detection experiment to illustrate the workings of our proposed mechanism and its ability to find solutions that go well beyond simple maximum-guided pose invariance, that overcomes the inherent ambiguity present in natural textures. From left to right: Input $x$ to the network; Gradient magnitude of $x$, as equivalent to max pooling over all rotations of the filter at each location; Pixel-wise predictions of the network $\mathcal{H}(x)$; Pose variables discovered by pose estimation network $\mathcal{F}(\Phi(x))$; Pixel-wise targets. I) Perfectly discovers the simple maximum-guided invariance rule to recover the target. II) More complex local relationships that can not be solved with simple invariants are recovered. Note, the two small blobs on the right, barely visible in input and gradient magnitude map, but still correctly segmented. III) Most challenging scenario, blobs and backgrounds can hardly be distinguished and the gradient magnitude does not contain much useful visible information for the task either. Note, that $\mathcal{F}(\Phi(x))$ learns an alternating pattern of rotation, that overcomes the irregular local gradients and recovers most of the targets successfully.
\label{fig:blobs}}
\vspace*{-3mm}
\end{figure}
In this experiment we empirically validate the effectiveness of our approach with a single dynamic steerable block by showing that it indeed learns adaptive and non-trivial invariants, that are conditioned on local neighborhoods in the image. To show this, we create an artificial dataset of random blobs, whose boundaries have to be detected by a dynamic steerable block as pixel-wise classification task. The dataset is infinite and created on the fly. In the first case, where blob and background are binary, this task can be solved with a simple gradient magnitude invariant and presents no challenge to our algorithm. 

A fully convolutional network baseline of the same size achieves almost the same performance. In the second and third example we sample textures from the KTH TIPS dataset~\cite{hayman2004significance} and fill blobs as well as background with different textures each. Here, gradient magnitude either only gives weak clues, or in many cases is unable to find any outline given by the target. In both cases, the fully convolutional baseline fails to converge. Remarkably, even though the dynamic steerable block does only receive two Gaussian derivative gradient filters as an input, it still manages to find highly non-linear steering patterns to recover the boundaries. The results are shown in figure~\ref{fig:blobs}. We show some unseen inputs alongside the manually calculated gradient magnitude, predictions and associated pose maps (only rotation variable shown) as estimated by the dynamic steerable block. Even in very hard cases, the dynamic block manages to largely recover the labeled boundaries. Results are evidence of the ability of our proposed method to learn adaptive invariants conditioned on the local context in the image in ways that go way beyond simple maximum response guided steering.

\subsubsection{Boundary Detection on BSD500}
\begin{figure}[H]
\includegraphics[width=.65\textwidth, center]{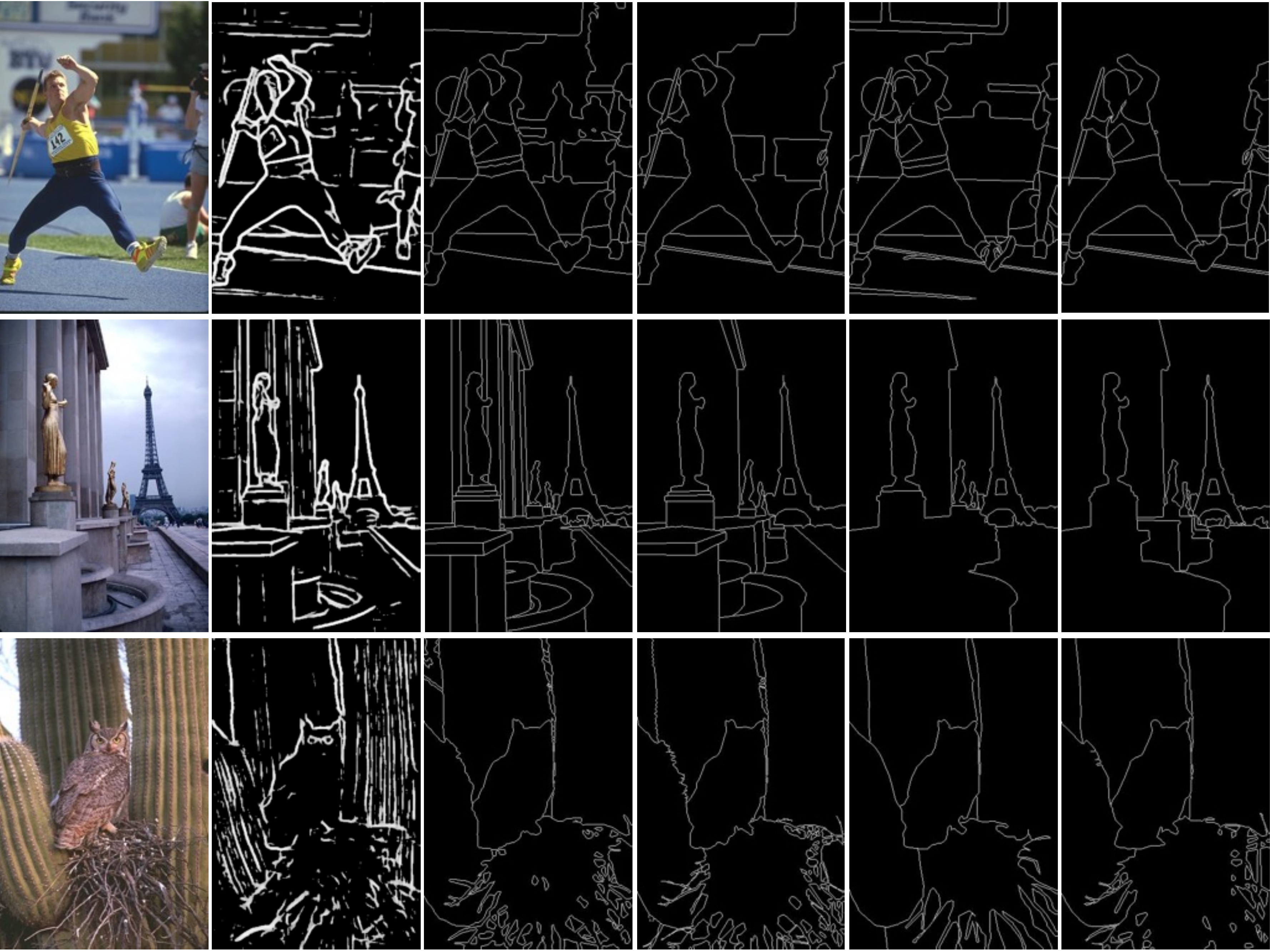}
\begin{center}
\begin{tabular}{@{}lccccc|c@{}} 
Method &  DynResNet &  H-Net & ResNet & Kivininen & HED & DCNN \\ 
 &  (Ours) & ~\cite{worrall2016harmonic}  &  (Ours-Static) &~\cite{worrall2016harmonic}  & ~\cite{worrall2016harmonic} & Pre-trained~\cite{kokkinos2015pushing} \\ \midrule
ODS & \textbf{0.732}   & 0.726  &0.720 &  0.702 & 0.697 &  0.813\\
OIS & \textbf{0.751} & 0.742  & 0.733 &0.715 & 0.709 &  0.831 \\
\end{tabular}
\end{center}
\caption{Results of non pre-trained state-of-the-art models alongside the currently best performing approach with pre-training on BSD500. We show results from our dynResNet, from left to right: Input image; Prediction and 4 different ground truth labels. Our dynamic steerable ResNet outperforms all other methods, including the same model with static blocks, when no pre-training is performed. We also outperform the fully rotation invariant H-Net in accordance to our findings from the texture experiments, where we found that stable, non-maximum guided invariance is likely superior in ambiguous cases. Note, that Kivinen and HED are non-pretrained re-implementations of \citep{kivinen2014visual,xie2015holistically} reported in \citep{worrall2016harmonic}. When pre-trained, HED and DCNN outperform non-pretrained approaches. For brevity we only report the state-of-the-art pre-trained DCNN.
\label{fig:bsd500}}
\end{figure}
In this experiment we apply the dynamic steerable blocks on a real task in which adaptive invariance is desirable. Recently, a fully equivariant CNN engineered to output locally rotation invariant predictions~\cite{worrall2016harmonic} performed very well on this task. 
We aim to show, that learned adaptive invariance is even more powerful, as even though final edge prediction is rotation invariant, intermediate processing benefits from a context dependent degree of invariance. We evaluate our method on the contour detection task of the Berkeley Segmentation Dataset. The dataset consists of 500 images divided in a train/val/test split. We tune hyperparameters on the validation set and report results on the test set, following the protocol in~\cite{bsds500}. Each image is labelled by 5-7 human annotators, resulting in multiple ground truth maps per image. We follow~\cite{worrall2016harmonic} and merge the different labels by majority vote into one. We minimize the pixelwise binary cross-entropy loss between ground truth and prediction and use class balancing because the classes are heavily imbalanced (many more background pixels than contour pixels). 

We use a ResNet model designed for segmentation~\cite{wu2016wider} and reduce its size drastically, due to the limited data-scenario we face. Eventually we use a network with the following structure: Conv2d[64]->DynResBlock[128]->StatResBlock[128]->DynResBlock[128]->StatResBlock[128]->Conv2D[256]. Thus, we alternate static and dynamic blocks as we found that this setup stabilizes training considerably on the validation set. The dynamic blocks are based on Gaussian derivatives and their exact setup is described in the appendix. We report OIS and ODS metrics as in~\cite{bsds500}, based on F-scores. Due to the subjective aspect of the contour segmentation, the task is ambiguous and the network implicitly has to estimate the level of detail at which to segment the boundaries. While in most cases the network's prediction agrees with at least one of the human annotators (first two rows of figure 5), it sometimes segments boundaries at a too high level of detail (last row of figure 5). Our results show that we outperform state-of-the-art approaches when they are not pre-trained on Imagenet, according to re-implementations from~\cite{worrall2016harmonic}, including our static ResNet baseline and the explicitly rotation invariant Harmonic Networks approach~\cite{worrall2016harmonic}. However, when pre-trained on Imagenet, DCNN (ODS: 0.813, OIS: 0.831) \citep{kokkinos2015pushing} and Holistically Nested Edge Detection (ODS: 0.782, OIS: 0.804) \citep{xie2015holistically} clearly outperform the non-pre-trained approaches.

In conclusion, we show that a learned piece-wise invariance is superior compared to a hand-crafted maximum response invariance as applied in the Harmonic Networks, supporting our findings from the texture experiments. However, semantic high-level knowledge other models acquire through Imagenet pre-training still seems to include forms of knowledge our geometrical regularization can not overcome, making our results especially interesting in domains where no large datasets for pre-training are available (e.g. medical imaging).

\section{Conclusion}
In summary, we have introduced a framework that opens up a range of novel and efficient ways to incorporate geometrical priors and regularization into deep networks, without restricting their theoretical expressiveness.

We introduced the notion of frame-based CNNs and derived dynamic steerable blocks from frame-based residual networks. In summary, we have shown that CNNs based on frames specifically designed for the data at hand facilitate optimization consistently even when data is abundant. On the other hand, our proposed Dynamic Steerable Blocks achieve superior performance when data is limited and are superior when filters that precisely and dynamically adapt to local patterns in non-trivial ways are needed.

\bibliography{egbib}
\end{document}